\documentclass{article}

\usepackage{arxiv}

\usepackage[utf8]{inputenc} 
\usepackage[T1]{fontenc}    
\usepackage{hyperref}       
\usepackage{url}            
\usepackage{booktabs}       
\usepackage{amsfonts}       
\usepackage{nicefrac}       
\usepackage{microtype}      
\usepackage{lipsum}
\usepackage{graphicx}
\graphicspath{ {./images/} }

\usepackage{amsmath,epstopdf,flushend,amssymb} 
\usepackage[figuresright]{rotating}
\usepackage{multirow,multicol,mathtools,comment}
\usepackage{xcolor}
\usepackage{pifont}
\newcommand{\cmark}{\ding{52}}%
\newcommand{\xmark}{\ding{56}}%
\newcommand{\greencmark}{{\color{green}\cmark}}
\newcommand{\redxmark}{{\color{red}\xmark}}

\newcommand{\zzie}{\emph{i.e.}~}
\newcommand{\zzeg}{\emph{e.g.}~}
\newcommand{\zzaka}{\emph{a.k.a.}~}

\newcommand*{\1}{\textcolor{magenta}} 
\usepackage{subfigure}

\title{Curvature-based Feature Selection with Application in Classifying Electronic Health Records}
  
\author{
    Zheming Zuo\thanks{Equal contribution.}\hspace{0.15cm}\thanks{Corresponding author.}\\
  Department of Computer Science\\
  Durham University\\
  Durham DH1 3LE, UK\\
  \texttt{zheming.zuo@durham.ac.uk} \\
  \And
 Jie Li$^{*}$ \\
  School of Computing, Engineering \& Digital Technologies\\
  Teesside University\\
  Middlesbrough TS3 6DR, UK\\
  \texttt{jie.li@tees.ac.uk} \\
  \And
 Han Xu \\
  Institute of Microelectronics, Chinese Academy of Sciences\\
  University of Chinese Academy of Sciences\\
  Beijing 100029, China\\
  \texttt{ann.gong.qifeng@gmail.com} \\
  \And
  Noura Al Moubayed\\
  Department of Computer Science\\
  Durham University\\
  Durham DH1 3LE, UK\\
  \texttt{noura.al-moubayed@durham.ac.uk} \\
}

\begin{document}
\maketitle
\begin{abstract}
Disruptive technologies provides unparalleled opportunities to contribute to the identifications of many aspects in pervasive healthcare, from the adoption of the Internet of Things through to Machine Learning (ML) techniques. As a powerful tool, ML has been widely applied in patient-centric healthcare solutions. To further improve the quality of patient care, Electronic Health Records (EHRs) are commonly adopted in healthcare facilities for analysis. It is a crucial task to apply AI and ML to analyse those EHRs for prediction and diagnostics due to their highly unstructured, unbalanced, incomplete, and high-dimensional nature. Dimensionality reduction is a common data preprocessing technique to cope with high-dimensional EHR data, which aims to reduce the number of features of EHR representation while improving the performance of the subsequent data analysis, \zzeg classification. In this work, an efficient filter-based feature selection method, namely Curvature-based Feature Selection (CFS), is presented. The proposed CFS applied the concept of Menger Curvature to rank the weights of all features in the given data set. The performance of the proposed CFS has been evaluated in four well-known EHR data sets, including Cervical Cancer Risk Factors (CCRFDS), Breast Cancer Coimbra (BCCDS), Breast Tissue (BTDS), and Diabetic Retinopathy Debrecen (DRDDS). The experimental results show that the proposed CFS achieved state-of-the-art performance on the above data sets against conventional PCA and other most recent approaches. The source code of the proposed approach is publicly available at \href{https://github.com/zhemingzuo/CFS}{\texttt{\1{https://github.com/zhemingzuo/CFS}}}.
\end{abstract}


	\section{Introduction}
    Disruptive technologies (DTs), including Industry 4.0, Artificial Intelligence (AI), Machine Learning (ML), Big Data, the Internet of Things (IoT), Virtual Reality (VR), etc., are innovations that significantly alters the way that consumers, industries or businesses operate~\cite{abdel2020intelligent}. Among those disruptive technologies, AI and ML are powerful tools to make computers learn and mimic human's thinking, and thus to help analyse the huge amount of medical data to identify potential health issues, such as \cite{chang2018computational, chang2018data}. 
	
	Due to the era of big data, large amounts of high-dimensional data have become available in a variety of domains, especially within the realm of digital healthcare~\cite{8370732}. The dramatically increased data volume has become a challenge for effective and efficient data analysis, as it significantly increases the memory storage requirement and computational costs~\cite{li2017feature}. To improve the quality of patient care more efficiently, Electronic Health Records (EHRs) are widely employed in the healthcare facilities for analysis. Currently, maintenance of such EHRs has become a crucial task in the medical sector. Patients' digital healthcare data is usually highly unstructured and consists of various features and diagnostics-related information. In addition, EHR data may include missing values and a certain degree of redundancy. Due to the incompleteness, imbalance, inherent heterogeneity, and high-dimensional features of EHRs, it is essential to apply disruptive technologies to provide effective and low-cost data analysis methods, and thus to explore such healthcare data for data mining and data analytics purposes. 
	
	Dimensionality reduction is an efficient data preprocessing technique for the analysis of high-dimensional data that aims to reduce the number of features while improving the classification performance (\zzeg treatment planning \cite{duanmu2020prediction}, survival analysis \cite{rietschel2018feature}, and risk prediction \cite{denaxas2018prediction}) and reducing the related computational cost~\cite{8361067}. It is important to identify the most significant factors that related to disease, which helps in removing unnecessary and redundant data from the given data sets, thus increasing the data analysis performance. The selection is usually achieved by either projecting the original data into a lower feature space, \zzaka feature extraction~\cite{han2018unified}, or selecting a subset of features from the original data, \zzie feature selection~\cite{liu2014feature}. For the latter, the least relevant features that are required to be removed can be identified by two criteria: a) features that are not correlated with any other features (\zzie redundancy); b) features that do not contribute to the classification decision (\zzie noise).
	
	
	It is well-known that the dimensionality reduction-based feature extraction approaches, such as Principal Component Analysis (PCA), reduces the number of features by mapping the original data set into a new feature space with lower dimensions, which changes or removes the physical meanings of original features. For instance, a low-complexity algorithm, which is developed based on Discrete Wavelet Transform (DWT), for electrocardiogram (ECG) data feature extraction, is proposed in \cite{mazomenos2013low}. In addition, to deal with the text classification problems, the Latent Semantic Index (LSI) \cite{dumais2004latent} and PCA algorithms have been employed to extract the text feature and have achieved good results \cite{levy2002least}. In contrast, methods of selecting a subset of features keep the physical meanings of the original features and enable models with better interpretability, but the underlying complexity of the subset evaluation metric may lead to unnecessary computational cost~\cite{JENSEN20151}. For example, a time series feature selection method is proposed to filter the available features with respect to their significance for the regression task \cite{christ2016distributed}. This motivates us to design an efficient selection-based feature selection method that could meet the requirement of the real-time system in the era of big data. 
	
	Recently, pervasive healthcare has become a central topic which has attracted intensive attention from academia, industry, as well as healthcare sectors~\cite{li2019bayesian,lu2020machine,livieris2019improving,tang2019construction,aydin2019construction,yang2018new,elyan2017genetic,apicella2019simple}. In this problem domain, highly class-imbalanced data sets with a large number of missing values are common problems~\cite{CervicalCancerDS2017}. It has been proved that the selected features might have a higher degree of usefulness in comparison with the projected features, due to preservation of the original semantics of the dimensions~\cite{8361067,liu2014feature}. Thus, we focus on selecting a sub-set of features, even using the anonymised data set (\zzeg one possible reason for having the missing attribute values could be that the participants or patients are reluctant to share personally identifiable information with the public \cite{CervicalCancerDS2017}), for efficient medical data classification.
	
	Based on the aforementioned two motivations, we address the issues of time complexity and efficiency in an intuitive, explainable fashion in this work. Our contribution is two-fold:
	
	
	\begin{enumerate}
		\item A filter-based feature selection method, called Curvature-based Feature Selection (CFS), is proposed to select discriminative attributes in line with the ranked and averaged curvature values for each dimension in the given EHR data set.
		
		\item The presented CFS approach is also embedded into a previously proposed fuzzy inference system, TSK+~\cite{Jie2017,li2018extended}, to constitute CFS-TSK+, for supporting better decision-making of clinical diagnosis, \zzie improving the performance of classifying digital healthcare data.
	\end{enumerate}
	
	The rest of the paper is structured as follows. Section~\ref{sec:bg} expresses the related works. Section~\ref{sec:appr} proposes our CFS approach and CFS-TSK+ classifier. Section~\ref{sec:exp} discusses the experimental results, and the paper is concluded in Section~\ref{sec:concl}.
	
	\section{Background}\label{sec:bg}
	In this section, the most recent developments of Machine Learning techniques in classifying medical data will be showcased first. This is followed by revisiting dimensionality reductions techniques for EHR data from the perspectives of feature extraction and feature selection, respectively.

	\subsection{Machine Learning for Digital Healthcare}
	In the past few decades, Machine Learning and deep learning algorithms have been widely proposed for solving healthcare problems, such as diagnosis prediction of various diseases including cervical cancer~\cite{ghoneim2020cervical}, breast cancer~\cite{devarriya2020unbalanced}, and thoracic disease~\cite{li2018thoracic}, which have usually taken the form of classification.
	
	Due to privacy considerations \cite{el2015anonymising,thompson2020ethical,chang2021ethical}, there is a large number of healthcare data sets containing missing values. To cope with this common issue, the Bayesian Possibilistic C-means (BPCM) \cite{li2019bayesian} was devised to interpolate the missing values by extending the Fuzzy C-Means clustering algorithm (to model the noise and uncertainty) with the support of Bayesian theory (to calculate cluster centroids). The Gene Sequence-based Auxiliary Model (GSAM)~\cite{lu2020machine}, as an ensemble learner, was proposed to predict the missing values via data correction and classify testing data samples via a combination of multiple weak learners within a gene auxiliary module.
	
	To enhance the classification performance in terms of accuracy, the Weight Constrained Neural Network (WCNN) was proposed \cite{livieris2019improving}. WCNN utilises network training to solve a constraint optimisation problem. An extension of the Broad Learning System (BLS) was devised by adding a label-based autoencoder (BLS II), to learn robust feature representations in an ensemble way, and also to tune the hyper-parameters in a Support Vector Machine (SVM), namely BLS II-SVM~\cite{tang2019construction}. Another ensemble learner, VIBES, was presented to detect the dependency between attributes in the given data set and to speed up the search for base learners~\cite{aydin2019construction}.
	
	In addition, Genetic Algorithm has been adopted to optimise the performance of Random Forest (RFGA)~\cite{elyan2017genetic}. Work towards the enhancement of activation functions in neural networks was also proposed, such as Variable Activation Function (VAF)~\cite{apicella2019simple} and Adaptive Takagi-Sugeno-Kang (AdaTSK)~\cite{8858838}. Apart from those adaptive action functions, a proposition of a two-layer mixture of factor analysers with joint factor loading (2L-MJFA) was presented to conduct the dimensionality reduction and classification together~\cite{yang2018new}. This is done by utilising two mixtures nested with each other, each of which containing several components, where each class of the data sets is represented in a specific mixture of factor analysers (MFA). Such an approach has been proven to be suitable for small-scale data sets, particularly those that contains a smaller number of data instances but includes a larger number of data attributes.
	
	\subsection{Dimensionality Reduction for EHRs}\label{sec:relatedworks}
\begin{figure}[!htbp]
	\centering
	\includegraphics[width=0.55\textwidth]{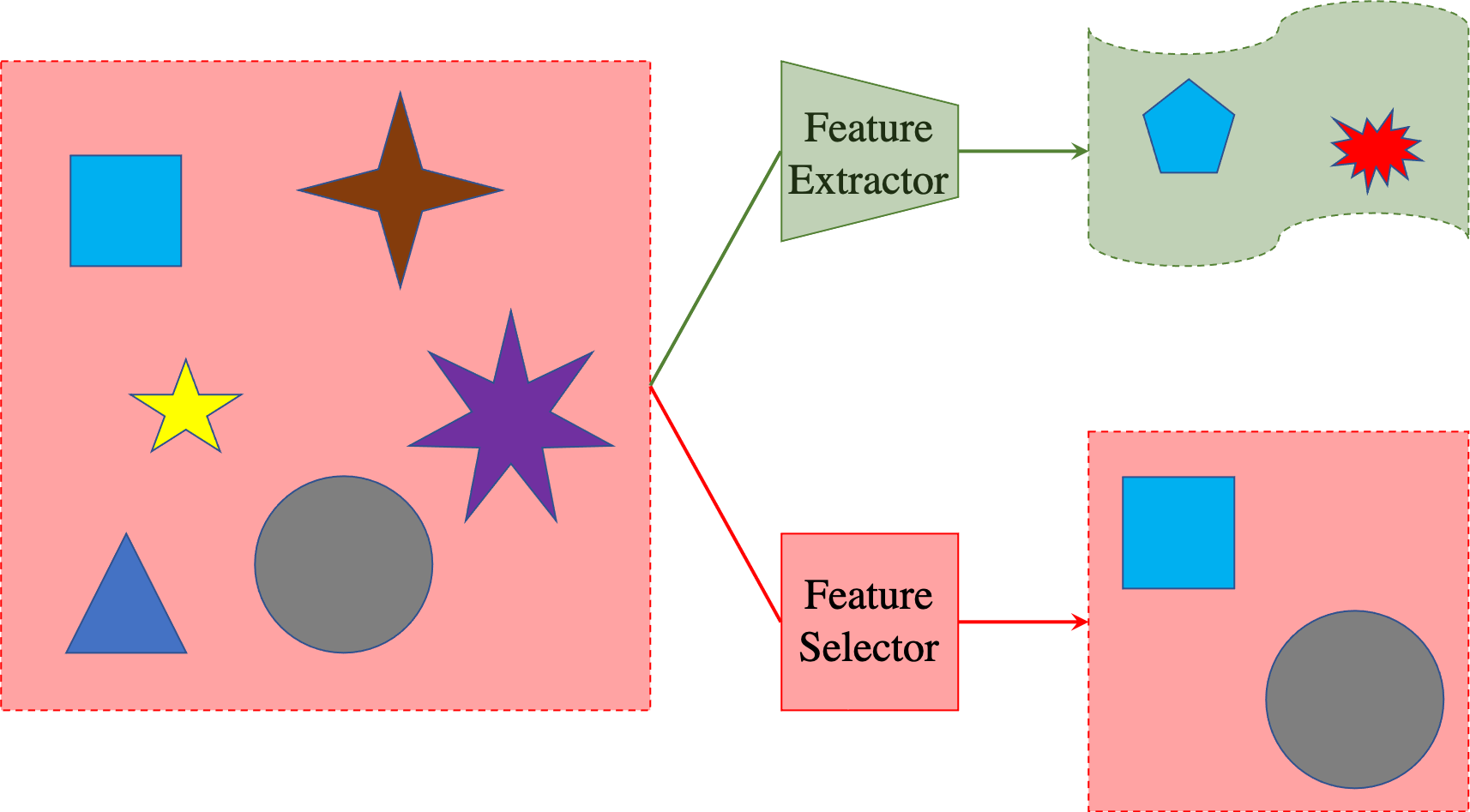}
	\caption{Illustration of feature extraction and feature selection.}\label{fig:fe-fs}
\end{figure}

	EHR data usually has a high-dimensions, thereby containing a large number of input features. It is noteworthy that some of the input features may not be relevant to the problem to be resolved. To effectively deal with such high-dimensional data, a typical solution is to apply specific techniques to reduce the dimensions of the original data set. Fundamentally, the dimensionality reduction techniques are typically divided into two aspects: 1) feature extraction, which combines the original features and creates a new set of feature representation; and 2) feature selection, which selects a subset of the original features~\cite{remeseiro2019review}. Fig.~\ref{fig:fe-fs} depicts the major difference between those two types of techniques, and both technologies are described below.  
	
	
	\subsubsection{Feature Extraction}\label{sec:fe}
	Feature extraction (FE), also termed as Feature Construction, is a substitute for feature selection that transforms the original data from a high-dimensional space into a low-dimensional one, as illustrated in the upper pathway of Fig.~\ref{fig:fe-fs}. By adopting this type of techniques, the problem is represented in a more discriminating (\zzie informative) space, thus leading to a more efficient analysis process. Such techniques have typically been applied in the fields of medical image analysis, such as Magnetic Resonance Imaging (MRI), Computed Tomography (CT) scans, Ultrasound and X-Rays~\cite{remeseiro2019review}. The common feature extraction techniques can be grouped into two main types: linear and non-linear. Linear feature extraction approaches, such as PCA, adopt the matrix factorisation method to transform the original data into a lower-dimensional subspace. For instance, PCA looks for ``principal components'' in the given data that are uncorrelated eigenvectors by considering the covariance matrix and its eigenvalues and eigenvectors~\cite{zuo2018gaze}. Although unsupervised PCA is highly effective in identifying important features of the data, it cannot easily determine the non-linear relationship among the features, which commonly exists in complex EHRs, especially, electrocardiogram (ECG), electroencephalography (EEG)~\cite{al2017enhanced}, and biological data~\cite{remeseiro2019review}. 
	
	Compared with linear feature extraction methods, which linearly map the original data into a low-dimensional subspace, non-linear feature extraction approaches work in different ways to represent the non-linear distribution, \zzeg Kernel PCA~\cite{aziz2017dimension}, Locally Linear Embedding (LLE)~\cite{aziz2017dimension}, and Self-Organising Maps (SOM)~\cite{li2019machine}. Such approaches worked based on the hypothesis that the data lies on an embedded non-linear manifold that has a lower dimension than the raw data space that lies within it~\cite{aziz2017dimension}. 
	
	Although the extracted features have a higher discriminating power that not only reduces the computational cost but also increases the classification accuracy, the combinations of the newly created sets of attributes may not have real meaning; therefore, those types of approaches may not be good methods with respect to readability, explainability, as well as transparency~\cite{remeseiro2019review}. 
	
	\subsubsection{Feature Selection}\label{sec:fs}
	Feature selection (FS) is a process of selecting a subset of the most important/relevant attributes from the given data set to use for model construction (\zzie data modelling). Similar to FE, the aim of FS is also to aid in the task of generating accurate predictive models; however, this is achieved by identifying and removing irrelevant and redundant features from the given data set which do not contribute to increasing the performance of a system model or, perhaps, may reduce the accuracy of the predictive model \cite{zuo2018grooming}, as depicted in the lower pathway of Fig.~\ref{fig:fe-fs}. Therefore, it is perfect when interpretability and knowledge extraction are crucial, \zzeg in medicine. Essentially, FS methods assess and evaluate the individual features in the original data set to determine the relevance of each feature for the given problem, so as to select the most relevant ones. In general, based on the relationship with the different learning methods, the process of feature selection can be categorised into three types: filter method, wrapper method, and embedded method \cite{liang2017text}. 
	
	\begin{itemize}
		\item \textit{Filter}: The filter method focuses on the general characteristics of each feature and ranks features based on a certain number of evaluation criteria. This is followed by a threshold value selection process in order to eliminate the features that are less than the selected crisp value. This method is computationally efficient and learning invariant, as it is independent of any learning algorithm. The limitation of such approaches is that there is no interaction between the classifiers, class labels (outputs), and dependency of one feature over others. Consequently, those approaches may fail to determine the most ``useful'' features \cite{liang2017text}. Typical filter feature selection methods include Information Gain (IG) \cite{azhagusundari2013ig}, Mutual Information (MI) \cite{amiri2011mi, pohjalainen2015feature}, and Chi-Square Test (CST) \cite{saengsiri2010cst}.
		\item \textit{Wrapper}: Unlike the filter method, the wrapper method depends on the performance of the learning algorithm to select features. In this method, candidate subsets of features are evaluated by an induction algorithm. The learning algorithms are employed to analyse the relationship between input features and the outputs (\zzie class labels), and thus identify the most useful/relevant features. Compared with filter methods, which are not computationally intensive, wrapper approaches usually have a complex progress and are more computationally costly than filter methods. In addition, this method is more prone to over-fitting on small training data sets \cite{liang2017text}.
		\item \textit{Embedded}: Though embedded method-based approaches still interact with learning algorithms for selecting relevant features, they conduct a different procedure from the filter and wrapper methods. In general, the embedded approaches can be described as a combination of the filter method and the wrapper method. They not only measure the relations between one input feature and its output feature (\zzie class labels) but also considers each feature's general characteristic itself locally for better local discrimination~\cite{ang2015supervised}. In particular, the embedded approaches firstly use the independent criteria to determine the optimal feature subsets from the given data set, and then, the learning algorithm is applied to finalise the final optimal feature subsets from the previous results. Compared with the wrapper method, the embedded approaches require low computational cost, and the chance of over-fitting is also reduced~\cite{ang2015supervised}. 
	\end{itemize}
	
	Recently, a hybrid method has also widely been employed to preprocess EHRs, in order to increase the model prediction capability. This method aggregates one or more FS approaches together, for example, filter and wrapper methods combined, to take the advantages of different methods, and hence to generate optimal results. The hybrid method can usually achieve better performance, \zzeg higher prediction accuracy; however, it also requires a higher computational cost~\cite{jain2018feature}. 
	
	\section{Proposed System} \label{sec:appr}
\begin{figure*}[ht]
		\centering
		\includegraphics[width=0.95\textwidth]{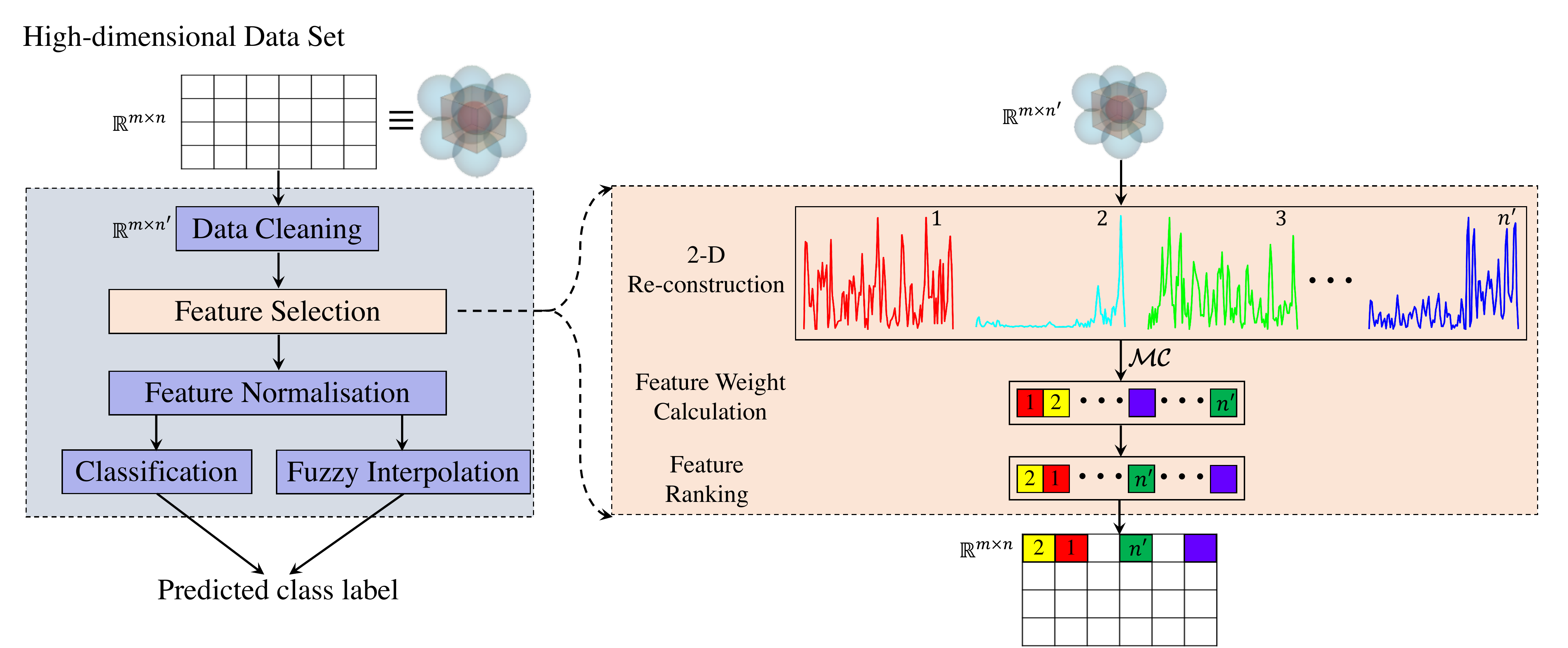}
		\caption{Architecture of the proposed CFS method.}\label{fig:pip}
	\end{figure*}

	A novel filter approach feature selection method, called Curvature -based Feature Selection (CFS), is proposed and detailed in this section. The system pipeline is illustrated in Fig.~\ref{fig:pip}, which comprises three main components: two-dimensional (2-D) data re-construction, feature weight calculation by Menger Curvature (depicted in Fig.~\ref{fig:mengercurvature}), and feature ranking. 
	
	\subsection{Menger Curvature}\label{sec:RoC}
\begin{figure}[!ht]
		\centering
		\includegraphics[width=0.52\textwidth]{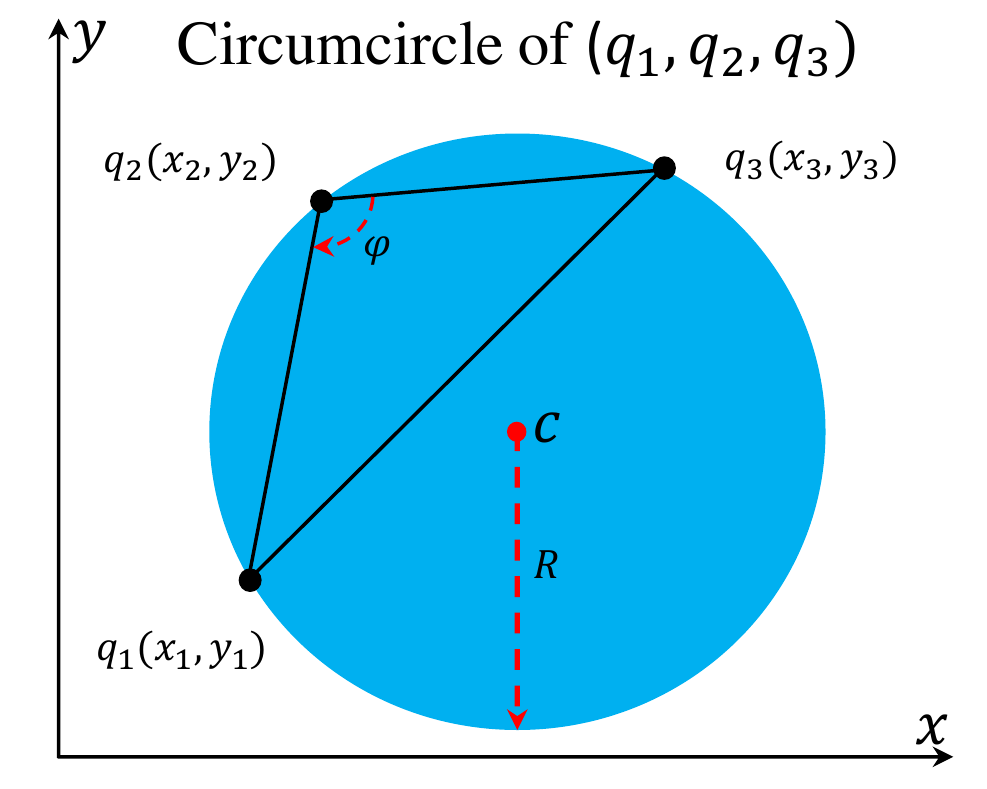}
		\caption{The measurement of the Menger Curvature on a 2-D space.}
		\label{fig:mengercurvature}
	\end{figure}

	Menger Curvature ($\mathcal{MC}$)~\cite{leger1999menger} measures the curvature of triple data points within the $n$-dimensional Euclidean space $\mathbb{E}^n$ represented by the reciprocal of the radius of the circle that passes through the three points $q_1$, $q_2$, and $q_3$ in Fig.~\ref{fig:mengercurvature}.
	
	In this work, only two-dimensional plane curves problems are considered. Given that $q_1(x_1,y_1),q_2(x_2,y_2)$, and $q_3(x_3,y_3)$ are the three points in a 2-D space, $\mathbb{E}^2$ and $q_1,q_2,q_3$ are non-collinear, as expressed in Fig.~\ref{fig:mengercurvature}, and $\mathcal{MC}$ on $B$ is calculated as:
	
	\begin{equation}\label{eq:MCvalue}
		\mathcal{MC}	(q_1,q_2,q_3) = \frac{1}{R} = \frac{2\textrm{sin}(\varphi)}{\|q_1,q_3\|},
	\end{equation}
	where $R$ represents the radius, $||q_1,q_3||$ denotes the Euclidean distance between $q_1$ and $q_3$, and  $\varphi$ is the angle of the $q_2$-corner of the triangle spanned by $q_1,q_2,q_3$, which can be calculated in line with the Law of Cosines:
	\begin{equation}
		\textrm{cos}(\varphi) = \frac{\|q_1,q_2\|^2 + \|q_2,q_3\|^2 - \|q_1,q_3\|^2}{2\cdot\|q_1,q_2\|^2\cdot\|q_2,q_3\|^2}.
	\end{equation}
	$\mathcal{MC}$ on points $q_1$ and $q_3$ is not calculable, as these points are boundary points. The efficacy of $\mathcal{MC}$ is confirmed in constructing a Mamdani fuzzy rule base~\cite{zuo2020CSRBG}.
	
	\subsection{Curvature-based Feature Selection}\label{sec:CFS}
	Assume that a high-dimensional raw data set, denoted as $\mathcal{U} \in \mathbb{R}^{m \times n}$, contains $m$ data instances, $n$ input attributes, and a single output feature $y$. In the real-world problem domain, a data cleaning process (\zzeg removing attributes with missing values) and a data normalisation phase (\zzeg bounding all the values within the interval of $[0,1]$) may be applied on $\mathcal{U}$ to obtain $\mathcal{U}^{'} \in \mathbb{R}^{m \times n^{'}}$ s.t. $n^{'} < n$. In this work, we adopt the Min-Max (MM) normalisation technique:
	
	\begin{equation}
		\mathcal{U}^{'} = \frac{\mathcal{U} - \textrm{min}(\mathcal{U})}{\textrm{max}(\mathcal{U})-\textrm{min}(\mathcal{U})}.
	\end{equation}

	This operation helps to cancel out the influence of possible large variations in the raw data set and guarantees that our CFS is able to compare the curvatures for each attribute in an equitable manner. In other words, all the attribute values are normalised to the same frame of reference to ensure the correct rankings are generated by CFS. The proposed CFS method is described as follows:
	
	\textbf{Step 1 -- \textit{2-D Data Re-construction}}: The first step of the proposed CFS is to break down the cleaned high-dimensional data set $\mathcal{U}^{'}$ into $n^{'}$ 2-D planes, which is implemented by combining all input attributes, $\mathcal{F}_i^{'}$ ($ 1\leqslant i \leqslant n^{'}$), and the output $y$. Thus, $\mathcal{U}^{'}$ can be decomposed to $n^{'}$ 2-D planes, represented as $\mathcal{P}_{(\mathcal{F}_i^{'},y)}$. 
	
	\textbf{Step 2 -- \textit{Feature Weighting}}: For each decomposed 2-D plane, $\mathcal{P}_{(\mathcal{F}_i^{'},y)}$, the Menger Curvature method, introduced in Section~\ref{sec:RoC}, is adopted to obtain the averaged curvature value of the feature $\mathcal{F}_i^{'}$. Given that a decomposed 2-D panel ($\mathcal{P}_{(\mathcal{F}_i^{'},y)}$) contains $m$ data instances, the Menger Curvature value ($\mathcal{MC}_{m_j}^i$) of data point $m_j$ $(2\leqslant j \leqslant m-1)$ can be determined by Eq.~(\ref{eq:MCvalue}). To this end, the mean of $\mathcal{MC}$ for $\mathcal{F}_i^{'}$, denoted as $\widehat{\mathcal{MC}_{\mathcal{F}_i^{'}}}$, is computed as:
	\begin{equation}
		\widehat{\mathcal{MC}_{\mathcal{F}_i^{'}}} = \frac{1}{m-2}\sum_{j=2}^{m-1} \mathcal{MC}_{m_j}^i,
	\end{equation}
	where $\mathcal{MC}_{m_j}^i$ represents the curvature value of the $m_j^{th}$ data point in feature $\mathcal{F}_i^{'}$. $\widehat{\mathcal{MC}_{\mathcal{F}_i^{'}}}$ indicates the corresponding weight of the feature $\mathcal{F}_i^{'}$, the greater value of $\widehat{\mathcal{MC}_{\mathcal{F}_i^{'}}}$ signifies a higher degree of importance of the corresponding feature $\mathcal{F}_i^{'}$ for the data set $\mathcal{U}^{'}$, and vice versa. 
	
	It is noted that for a given data set, changing the order of data points leads to a different value of $\mathcal{MC}_{m_j}^i$. However, the order of the all data instances will be changed accordingly. For this reason, the ranking of the features remain the same.
	
	\textbf{Step 3 -- \textit{Feature Ranking and Feature Selection}}: A conventional ordinal ranking method is used to rank the features, based on the obtained $\widehat{\mathcal{MC}_{\mathcal{F}_i^{'}}}$. Therefore, the features of $\mathcal{U}^{'}$ are ranked. This is followed by selecting the corresponding features from the raw data set $\mathcal{U}$. Given a threshold $\partial$, the features with $\widehat{\mathcal{MC}_{\mathcal{F}_i^{'}}}$ greater than the given threshold $\partial$ will be selected. Equivalently, a Top$K$ method can be employed:
	\begin{equation}\label{eq:topK}
		\mathcal{U}^{''} \vcentcolon= \mathcal{U} \bigg[ \textrm{Rank}^{\textrm{Top$K$}}\Big( \widehat{\mathcal{MC}_{\mathcal{F}_i^{'}}}\Big)\bigg],
	\end{equation}
	such that $\mathcal{U}^{''} \in \mathbb{R}^{m\times n^{'}}$. To this end, we have reduced the dimensionality of $\mathcal{U}$ to $\mathcal{U}^{''}$ while preserving the statistical nature of the original data set. Then, in line with the rest of the parts shown in Fig.~\ref{fig:pip}, the obtained $\mathcal{U}^{''}$ will be further normalised and classified.
	
	\subsection{Feature Normalisation} \label{sec:featNorm}
	In this work, we involve two different data normalisation processes: one before curvature calculation, the other after. The first data normalisation process is applied to make sure that the curvatures for each attribute can be compared equitably. The second data normalisation process is optional. It is used to reduce the noise and hence improve the classification performance on EHR data sets. 
	
	Concretely, the Min-Max data normalisation technique has been adopted for feature ranking, which aims to ensure each data attribute is compared in an equitable manner. To improve the performance of classification and ensure the degree of membership in the TSK+ is calculable,
	the selected features in $\mathcal{U}^{''}$ are further normalised using a total number of eight normalisation techniques~\cite{8858838} in this work, including Min-Max (MM) normalisation, $\ell$1-normalisation, $\ell$2-normalisation, Power Normalisation (PN), and its variants (\zzie $\ell$1PN, $\ell$2PN, PN$\ell$1, and PN$\ell$2).
	
	\subsection{Classification} \label{sec:featClasf}
	To classify the selected and normalised features, nine classifiers~\cite{zuo2018grooming,8858838} are used, namely Gaussian Na\"ive Bayes (GNB), Random Forest (RF), AdaBoost (AB), Logistic Regression (LR), Linear Support Vector Machine (Linear SVM), Decision Tree (DT), $k$ Nearest Neighbours ($k$NN), and Back-Propagation Neural Network (BPNN). Additionally, we also combine the proposed CFS method with TSK+ (CFS-TSK+) and evaluate its performance for the classification of four benchmark medical data sets.
	
	\section{Experiments} \label{sec:exp}
	For performance verification and evaluation, the proposed CFS method is compared against PCA, IG, MI, and CST on four benchmark clinical data sets. In the following, we describe the data sets and the experimental setup we used to examine the aforementioned techniques.
	
	\subsection{Data sets}
\begin{figure}[!ht]
		\setlength{\tabcolsep}{-8pt}
		\centering
		\begin{tabular}{cc}
			\hspace{-0.3em}\includegraphics[width=0.54\linewidth]{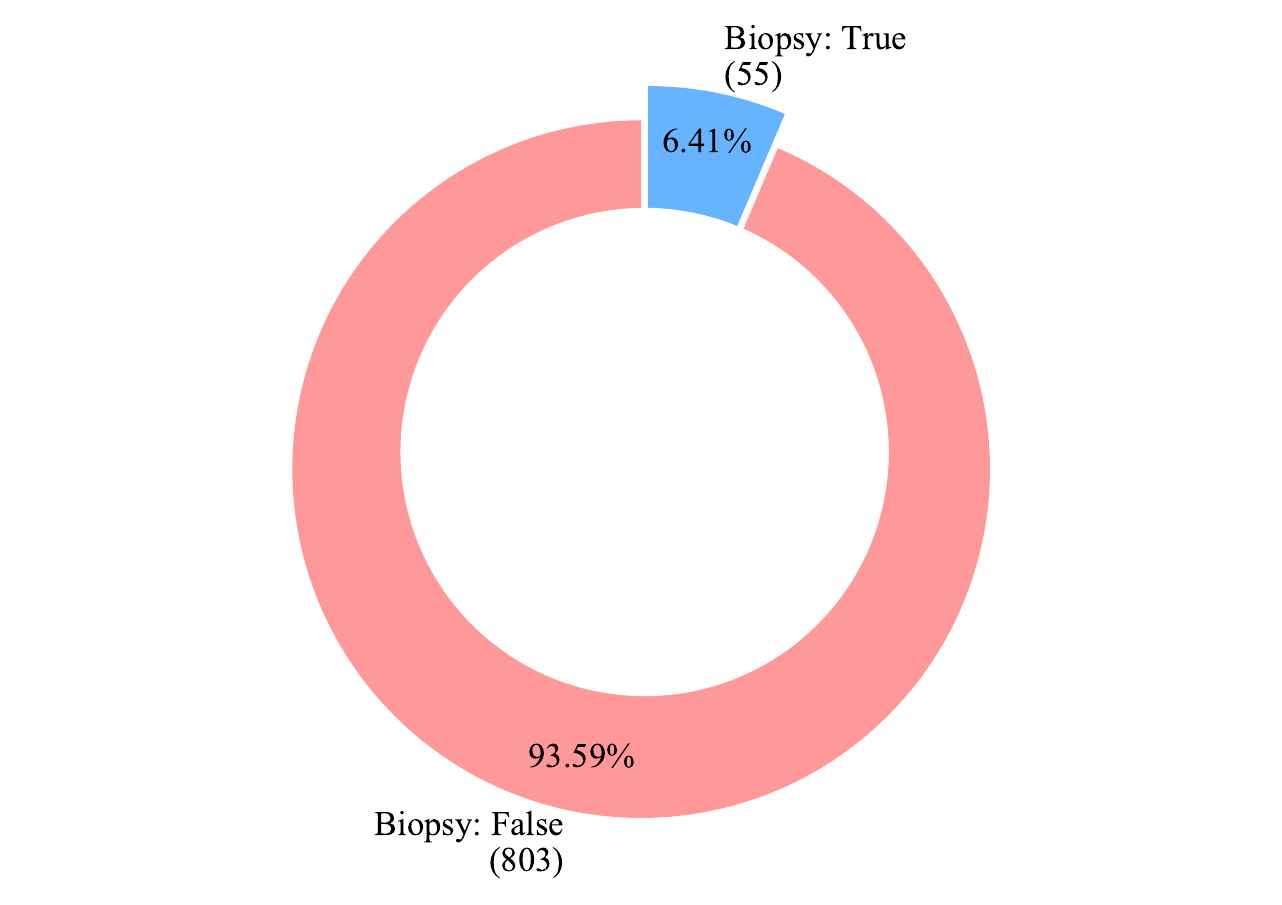} & \includegraphics[width=0.54\linewidth]{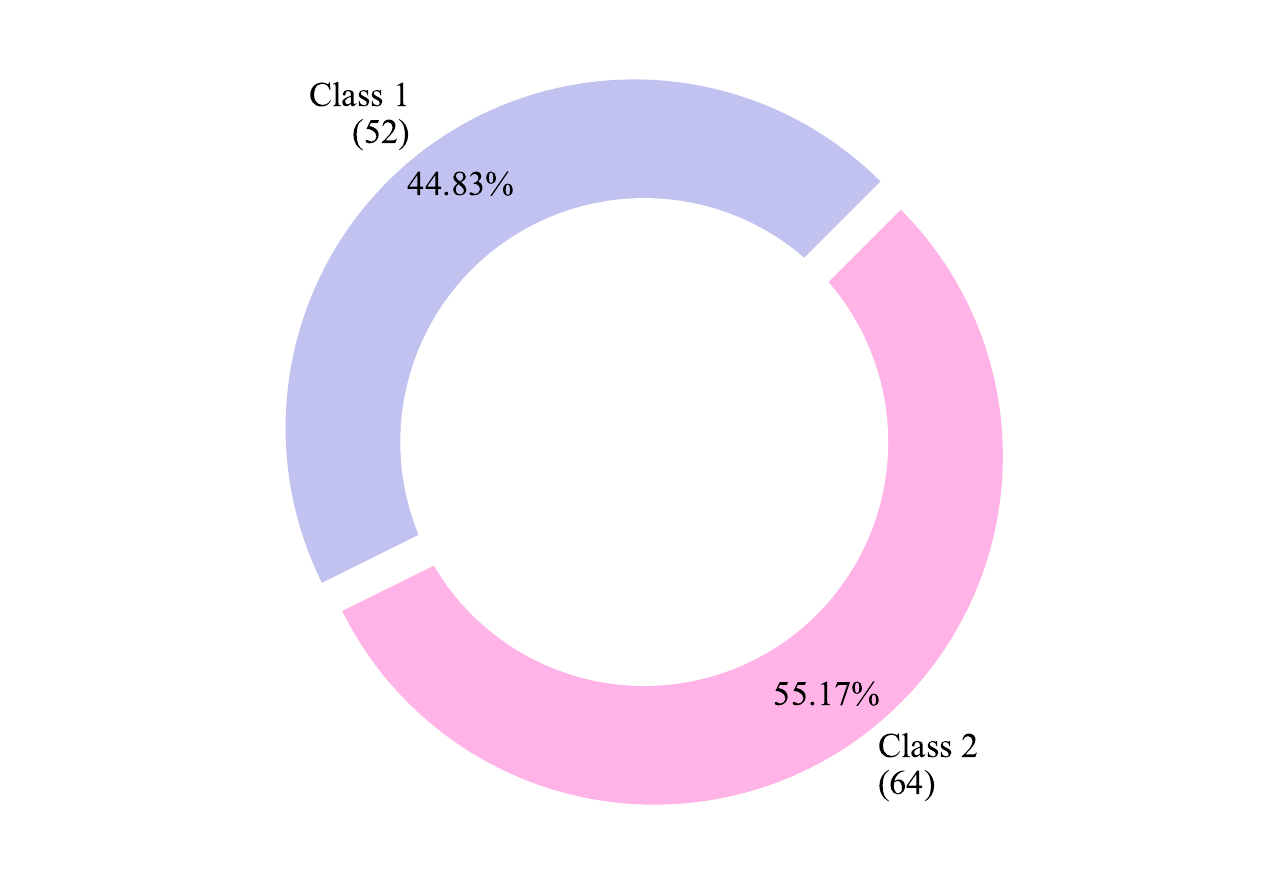}\\
			(a) CCRFDS & (b) BCCDS\\
			\hspace{-0.3em}\includegraphics[width=0.54\linewidth]{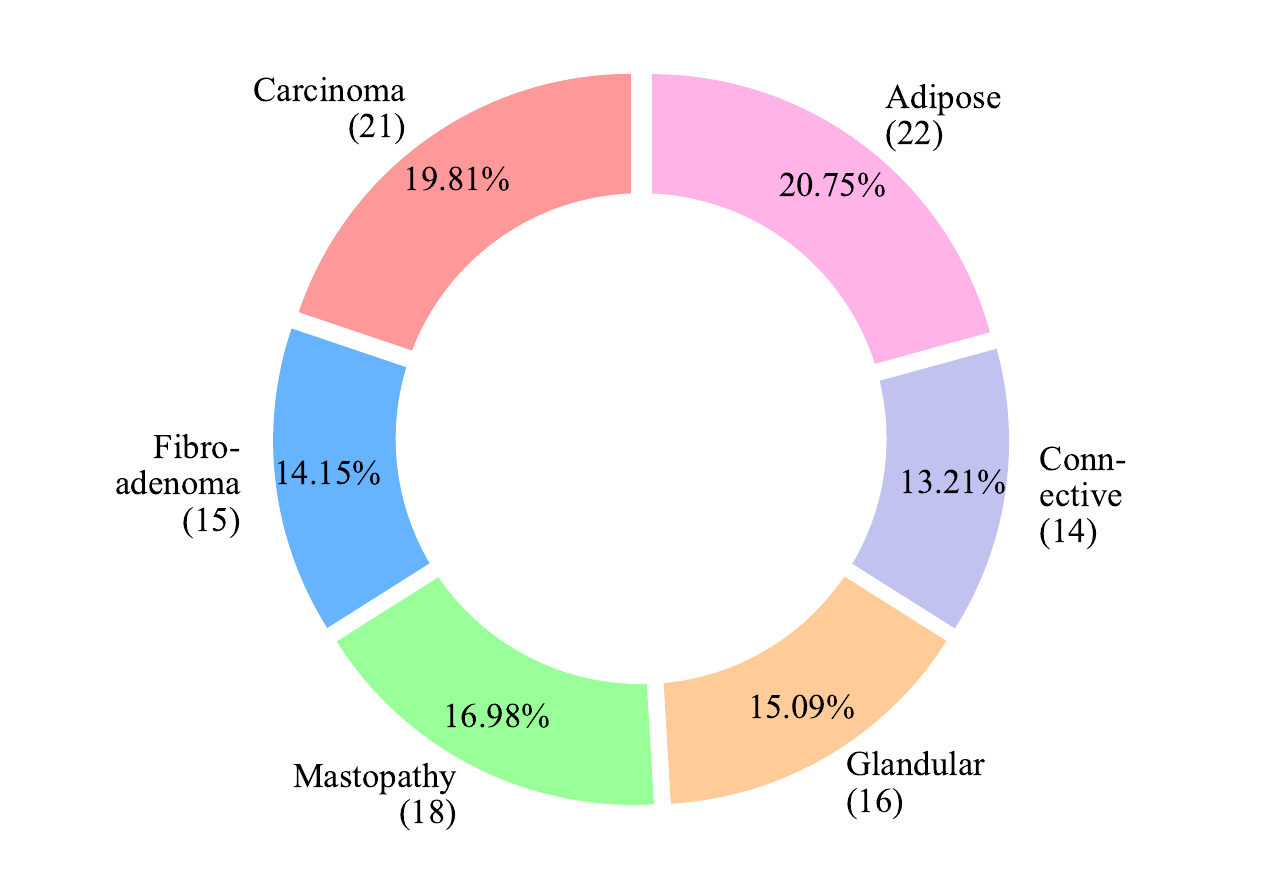} & \includegraphics[width=0.54\linewidth]{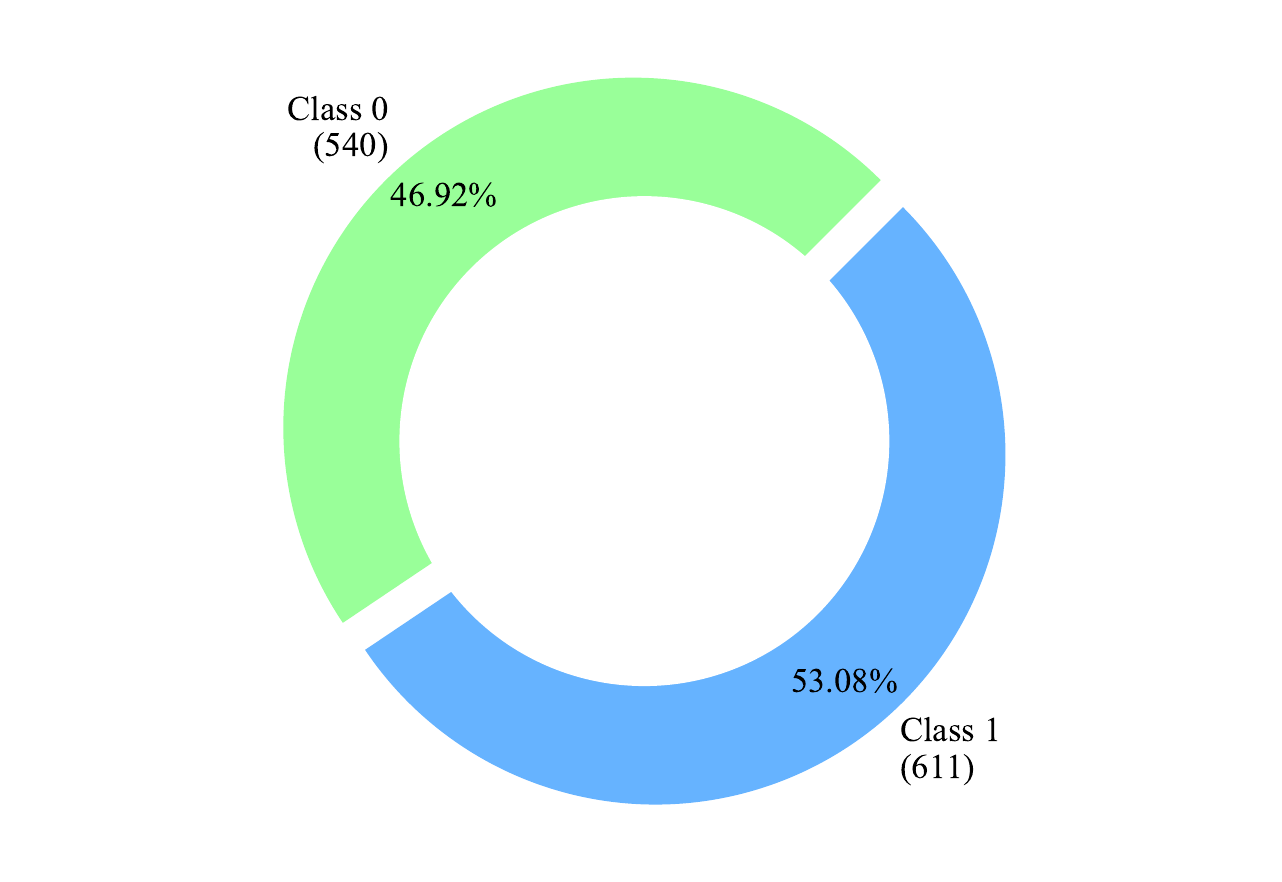}\\
			(c) BTDS & (d) DRDDS
		\end{tabular}
		\caption{Data instances percentage for each data set. Best viewed in colour.}
		\label{fig:DS}
	\end{figure}

	\begin{table}[!h]
		\centering
		\caption{Data set structures. $^\dagger$ denotes the exclusion of class label. $^\star$ represents the number of dimensions contain the missing values.}
		\begin{tabular}{l|cccccc}
			\hline
			\rotatebox{90}{Data Set}
			& \rotatebox{90}{\texttt{\#} of instances} & \rotatebox{90}{\texttt{\#} of dim.$^\dagger$}
			& \rotatebox{90}{\texttt{\#} of classes}
			& \rotatebox{90}{Missing values?}
			& \rotatebox{90}{\texttt{\#} of dim. $^\star\dagger$} 
			& \rotatebox{90}{Year published}\\
			\hline\hline
			
			CCRFDS & 858 & 35 & 2 & \greencmark & 26 & 2017\\ 
			\hline
			BCCDS & 116 & 9 & 2 & \redxmark & N/A & 2018\\ 
			\hline
			BTDS & 106 & 9 & 6 & \redxmark & N/A & 2010\\ 
			\hline
			DRDDS & 1,151 & 19 & 2 & \redxmark & N/A & 2014\\ 
			\hline
		\end{tabular}
		\label{tbl:ppmlComparison}
	\end{table}
	\textbf{Cervical Cancer (Risk Factors) Data Set}~\cite{CervicalCancerDS2017} (CCRFDS) comprises demographic information, habits, and historic medical records of 858 patients, with some missing attribute values due to the consents of the participating patients. The data set is categorised by the Boolean value of the biopsy. CCRFDS is also highly class-imbalanced, \zzie only 18 out of 858 participants have cancer, demonstrated in Fig.~\ref{fig:DS}(a). Concretely, without including the class label, the original CCRFDS consists of 858 instances, each of which containing 35 attributes. However, there are 26 attributes containing missing values `?' due to the purpose of privacy protection. Thus, without the inclusion of the class label, we use 858 instances, each of which containing 9 attributes.
	
	\textbf{Breast Cancer Coimbra Data Set}~\cite{BreastCancerCoimbraDS2018} (BCCDS) is composed of 116 data instances (each of which containing 9 attributes) that can be grouped into two categories, \zzie healthy controls, and patients.
	
	\textbf{Breast Tissue Data Set}~\cite{BeastTissueDS2010} (BTDS) contains 106 data instances, each of which with 9 feature dimensions that can be classified into six categories, including carcinoma, fibro-adenoma, mastopathy, glandular, connective, and adipose.
	
	\textbf{Diabetic Retinopathy Debrecen Data Set}~\cite{DiabeticRetinopathyDebrecenDS2014} (DRDDS) includes 1,151 data instances that are categorised into two classes which respectively indicate having and not having Diabetic Retinopathy (DR).
	
	\subsection{Dealing with Missing Data}\label{sec:missingvalues}
	Missing data, \zzaka missing values, is a common issue in the digital healthcare domain. As introduced above, missing data could reduce the statistical power of the predictive model, as well as lead to incorrect or invalid results. Therefore, an extra stage may be required for handling the issue of missing data. There are two major methods that could be adopted to cope with this issue, \zzie data imputation and case deletion~\cite{kang2013prevention}. Concretely, imputation is the process of inserting the missed values with substituting components. Several approaches have been well discussed in the literature, such as mean, Multiple Imputation with Chained Equations-Full (MICE-full), and missForset~\cite{luo2016using}. Among those methods, the mean imputation approach imputes missing values as the mean of the available values for each variable, MICE-full and missForset then use Machine Learning algorithms, \zzeg Random Forest, to predict the missing values based on the observed values of a data matrix. For the latter, all data instances with missing data are simply omitted/removed, and then only the remaining data is used for the analysis. Alternatively, the remaining data could be retained by removing the attributes in which the missing values are situated. In this work, we merely apply the attribute deletion method on CCRFDS.
	
	\subsection{Experiment Setup}
	
	Among all the four selected data sets, we perform data cleaning on CCRFDS, \zzie the attributes, which contain the missing value, are eliminated. Therefore, the CCRFDS used in this work contains 858 data instances, each of which having 9 attributes (not including the class label). For all the other three data sets, we use the originally published data.
	
	For selecting features, we compare the proposed CFS using the Top$K$ method defined in Eq.~(\ref{eq:topK}) with the well-known PCA, IG, MI, and CST by varying the numbers of the selected features for classification. That is, we select 7 out of 9 attributes (in CCRFDS, BCCDS, and BTDS) and 15 out of 19 attributes (in DRDDS). To normalise the selected features, 8 normalisation methods introduced in Section~\ref{sec:featNorm} are employed, in which the power coefficient in PN and its variants are set to 0.1. For classification, 9 classifiers (introduced in Section~\ref{sec:featClasf}) are employed with the configuration information: the maximum number of estimators is 100 at which the boosting is terminated in AB; the $\ell$1 regularisation is adopted as the penalty function in LR; the Gini index is employed as the criterion in DT and the maximum depth of the tree is valued as 5; the number of neurons in a single hidden layer of BPNN is set to 20; the $k$ is valued as 3 in $k$NN. The mean accuracy is reported for the nine employed classifiers for performance comparisons via the 10-Fold cross-validation.
	
	\subsection{The Efficacy of CFS}
\begin{figure*}[!ht]
		\centering
		\includegraphics[width=0.98\textwidth]{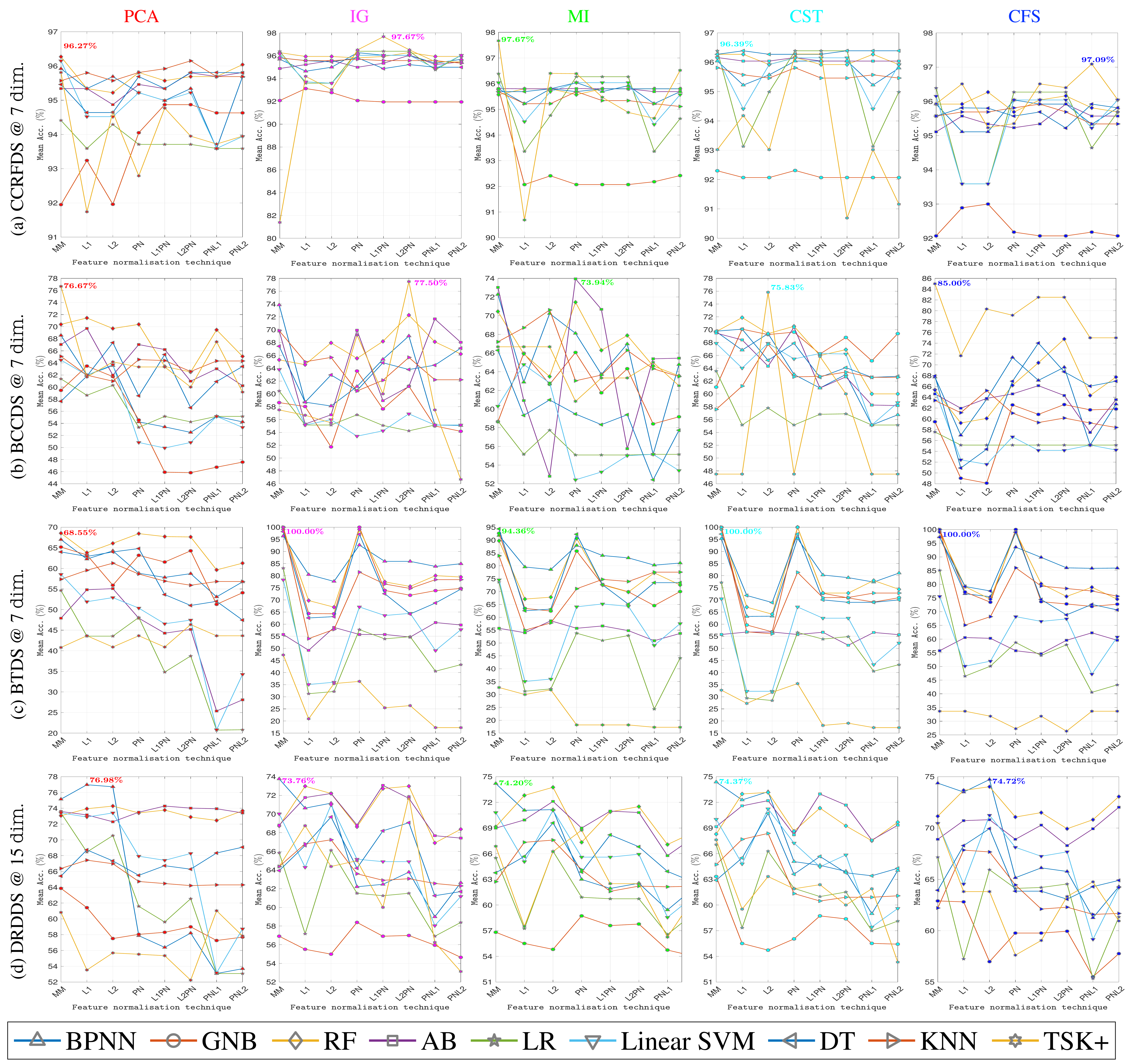}
		\caption{Detailed classification accuracies of PCA, IG, MI, CST, and CFS feature selection methods on four EHR data sets by varying both feature normalisation methods and classifiers. Best viewed in colour.}
		\label{fig:CFS}
	\end{figure*}

\begin{figure}[!ht]
		\centering
		\includegraphics[width=0.67\textwidth]{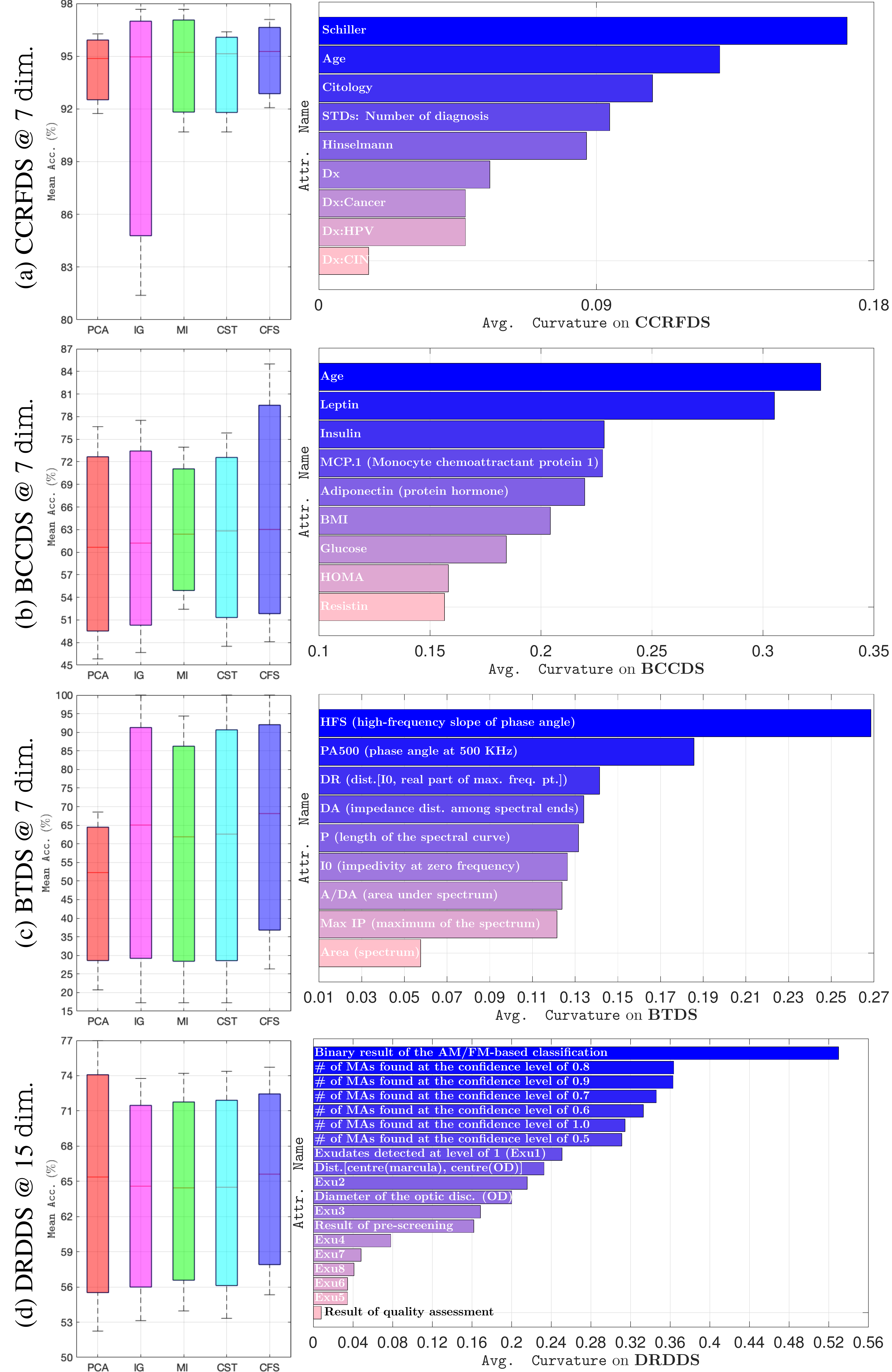}
		\caption{Classification summaries of PCA, IG, MI, CST, and CFS on four EHR data sets by varying both feature normalisation methods and classifiers. The left column summarises the classification statistics presented in Fig. \ref{fig:CFS}, and the right columns lists the feature rankings provided by CFS in a descending order over each data set for intuitive explainability. Best viewed in colour.}
		\label{fig:CFS2}
	\end{figure}

	The detailed comparisons among the CFS and four counterparts are shown in Fig. \ref{fig:CFS} with Top Mean Accuracy (TMA) highlighted using different colour codes. In addition, those accuracies are further summarised in the first column of Fig. \ref{fig:CFS2}. In general, CFS outperforms PCA, IG, MI, and CST on the four data sets. Concretely, CFS yields averaged mean accuracies of 95.27\%, 63.02\%, 68.15\%, and 65.61\%, versus 94.87\%, 60.65\%, 52.27\%, and 65.37\% resulting from PCA; 94.96\%, 61.20\%, 65.06\%, 64.59\% generated by IG; 95.22\%, 62.40\%, 61.90\%, and 64.44\% produced by MI; and 95.14\%, 62.82\%, 62.62\%, and 64.50\% yielded by CST. This observation indicates that the CFS is generally more competitive than PCA, IG, MI, and CST.
	
	For the CCRFDS, as depicted in Fig. \ref{fig:CFS}(a), the TMA yielded by the CFS-based, PCA-based, IG-based, MI-based, and CST-based are 97.09\%, 96.27\%, 97.67\%, 97.67\%, and 96.39\%, respectively. Among those best results, the MM normalisation approach is more suitable for PCA, MI, and CST, whereas the peak performance of CFS and IG are yielded by utilising PN$\ell$1 and $\ell$1PN normalisation approaches, respectively.
	
	
	
	For the BCCDS, as depicted in Fig. \ref{fig:CFS}(b), the TMA obtained by the CFS-based, PCA-based, IG-based, MI-based, and CST-based are 85.00\%, 76.67\%, 77.50\%, 73.94\%, and 75.83\%, respectively. The conventional data normalisation approach tends to be an appropriate choice for CFS (MM), PCA (MM), and CST ($\ell$2), whereas the variants of PN are more preferred by IG ($\ell$2PN) and MI ($\ell$1PN).

	
	
	For the BTDS, as depicted in Fig. \ref{fig:CFS}(c), we show that all feature selection methods achieve TMA using the MM normalisation approach. Furthermore, all the experimented filter feature selection methods (\zzie CFS, IG, MI, and CST) significantly outperform the feature extraction one (\zzie PCA). Among the feature selection approaches, our CFS is confirmed as having the highest degree of stability in terms of 68.15\% averaged mean accuracy. On this basis, CFS is capable of helping to differentiate between certain categories of healthy or pathological breast tissue.
	
	
	
	For the DRDDS, as depicted in Fig. \ref{fig:CFS}(d), TMA of all the experimented methods are all yielded by using conventional feature normalisation approaches, \zzie MM (IG, MI, and CST), $\ell$1 (PCA), and $\ell$2 (CFS). Specifically, in contrast to BTDS, PCA (76.98\%) achieves better TMA than all the filter feature selection methods, though on average CFS (65.61\%) is the best.

	
	
	To better explain the reason that CFS is a more competitive candidate of feature selection in comparison with PCA and the other three filter feature selection methods, we visualised the ranking of all the attributes (\zzie feature importance ranking) generated by CFS in the right column of Fig. \ref{fig:CFS2}.

	\subsection{The Efficiency of CFS-TSK+}
	Having the representative feature extraction (PCA) and feature selection (CFS) method, we detail the peak performances (TMA) of PCA-TSK+ and CFS-TSK+ in Table~\ref{tbl:cfstsk+}. Notably, in conjunction with Fig.~\ref{fig:CFS}, CFS-TSK+ achieved the best performance in the data sets of CCRFDS, BCCDS, and DRDDS. This observation confirmed the practicability and efficiency of combining the CFS with TSK+. 
	
	\begin{table}[!ht]
		\centering
		\caption{Optimal performance comparisons of PCA-TSK+ and CFS-TSK+. Rule base employed in PCA-TSK+ and CFS-TSK+ were not optimised. ($\cdots$) denotes multiple combinations achieving the same performance. Best performance is marked in bold.}
		\begin{tabular}{l|c|c}
			\hline
			Data Set
			& PCA-TSK+ (\%)$\uparrow$ & CFS-TSK+ (\%)$\uparrow$\\
			\hline
			\hline
			CCRFDS & 95.81 (MM) & \textbf{97.09} (PN$\ell$1)\\ 
			\hline
			BCCDS & 76.67 (MM) & \textbf{85.00} (MM)\\ 
			\hline
			BTDS & \textbf{46.36} ($\ell$2PN) & 33.65 ($\cdots$)\\ 
			\hline
			DRDDS & 61.03 (PN$\ell$1) & \textbf{70.48} (MM)\%\\ 
			\hline
		\end{tabular}
		\label{tbl:cfstsk+}
	\end{table}

	However, the best performances of CFS-TSK+ and PCA-TSK+ on BTDS help us to identify the possible drawback of the TSK+ in coping with classification tasks. That is, TSK+ is not sensitive to formulate a class boundary when the given data samples are sparsely distributed in the feature space. Alternatively, the rule base is not generalised well in the step of clustering where each cluster is corresponding to a fuzzy rule. Based on the time consumption, we did not perform rule base optimisation in this work as this was slightly beyond our scope. For the last data set, DRDDS, owing greatly to the lack of expert knowledge, it is not possible to explain how reasonable the ranked results of CFS in comparison with the rest of the data sets are, which are more common sensible. The task of designing a self-explainable component could be treated as active future work.

\begin{table}
	\centering
	\caption{Performance summary and comparisons. DS denotes data set and TMA represents Top \texttt{Mean Acc.}. Abbreviations of all the experimented data sets are summarised in Table \ref{tbl:ppmlComparison}. $\cdots$ represents multiple combinations achieved the same performance (see Fig. \ref{fig:CFS}(c). Best performance is marked in bold.}
	\begin{tabular}{c|c|c|c}
		\hline
		DS
		& Method & \texttt{Dim.}$\downarrow$ & TMA (\%)$\uparrow$\\
		\hline
		\hline
		\multirow{6}{*}{\rotatebox{90}{CCRFDS}} & BPCM+$5$NN \cite{li2019bayesian} & 9 & 80.00\\ 
		& GSAM \cite{lu2020machine} & 9 & 83.16\\
		& PCA-RF & 7 & 96.27\\ 
		& CST-DT & 7 & 96.39\\
		& CFS-TSK+ & 7 & 97.09\\
		& IG/MI-TSK+ & 7 & \textbf{97.67}\\
		
		\hline
		\multirow{7}{*}{\rotatebox{90}{BCCDS}} & WCNN \cite{livieris2019improving} & 9 & 62.00\\ 
		& BLS II-SVM \cite{tang2019construction} & 9 & 71.20\\
		& MI-AB & 7 & 73.94\\
		& CST-TSK+ & 7 & 75.83\\
		& PCA-TSK+ & 7 & 76.67\\
		& IG-TSK+ & 7 & 77.50\\
		& CFS-TSK+ & 7 & \textbf{85.00}\\
		
		\hline
		\multirow{5}{*}{\rotatebox{90}{BTDS}} & VIBES \cite{aydin2019construction} & 9 & 65.09\\
		& PCA-RF & 7 & 68.55\\
		& 2L-MJFA \cite{yang2018new} & 9 & 75.27\\
		& MI-DT & 7 & 94.36\\
		& CFS/IG/CST-$\cdots$ & 7 & \textbf{100.00}\\
		
		\hline
		\multirow{7}{*}{\rotatebox{90}{DRDDS}} & RFGA \cite{elyan2017genetic} & 19 & 68.26\\
		& VAF \cite{apicella2019simple} & 19 & 73.35\\
		& IG-BPNN & 15 & 73.76\\
		& MI-BPNN & 15 & 74.20\\
		& CST-BPNN & 15 & 74.37\\
		& CFS-BPNN & 15 & 74.72\\ 
		& PCA-BPNN & 15 & \textbf{76.98}\\ 
		\hline
	\end{tabular}
	\label{tbl:csktsk+}
\end{table}

	\subsection{Discussions}
	
	To summarise the proposed approach, we compare our CFS with PCA, IG, MI, CST and other recent competitive works in Table~\ref{tbl:csktsk+}. 
	
	Though CFS achieved the best performance among all the four medical data sets, and CFS-TSK+ yielded the two highest mean accuracies on two data sets, we identified that a possible drawback of the proposed CFS is the lack of better explainability when the domain (\zzeg clinical science) knowledge is not available. This might be mitigated by predicting the missing values on the anonymised data set and training a self-explainable component. In addition, a number of curvature-based methods, \zzeg Integral Menger Curvature (IMC) and Gaussian Curvature (GC), can also be adopted to determine the curvature values of given data sets. Among such methods, Gaussian Curvature is usually applied to analyse the curvature of folded surfaces \cite{richard1994detection}. And the Integral Menger Curvature is the integral of the Menger Curvature over the whole given surface, which is related to the surface's Euler characteristic. The proposed method considers the curvature value of three discrete points. For this reason, GC and IMC is not implemented; however, it is still worthwhile to investigate how GC and IMC can support this approach in the future. Finally, four EHR data sets have been employed for the system evaluations; more complex data sets, such as those including labelled noise, would be applied to investigate the performance of the proposed technique when dealing with more complex situations. 
	
	\section{Conclusion} \label{sec:concl}
	A novel filter-based feature selection method, CFS, has been presented as contributing to the classification performance of clinical (EHR) data sets. State-of-the-art performance on all four benchmark clinical data sets has been achieved by the proposed CFS approach. Though lacking expert knowledge of clinical science, we visualise the results of feature ranking given by CFS to support better explainability. It is noteworthy that the self-explainability of CFS and sparsity awareness of class boundaries of CFS-TSK+ are observed as possible future directions. In addition, a given threshold is required by the proposed CFS approach for selecting the ranked features. It would be worthwhile to investigate how this threshold can be intelligently determined in the future. Also, due to the limitation of the filter method, the selected features may not be relevant to the class labels. It is interesting to introduce some learning approaches, such as \cite{li2016experience}, to consider the relationship between features and class labels while ranking the features for the MLaaS \cite{chang2018overview}.
	
	\section*{Acknowledgements}
	This work was sponsored by the UK Research and Innovation fund (project 312409) and Cievert Ltd. 
	
	\bibliographystyle{unsrt}  
\bibliography{ref}

\end{document}